\documentclass[runningheads]{llncs}
\usepackage[T1]{fontenc}
\usepackage{graphicx}
\usepackage{amsmath,amssymb}
\usepackage{url}
\usepackage{booktabs}   
\usepackage{multirow}   

\DeclareMathOperator{\softmax}{softmax}
\DeclareMathOperator{\KL}{KL}

\usepackage{xcolor}
\usepackage{hyperref}
\hypersetup{
  colorlinks=true,
  linkcolor=blue,  
  citecolor=blue,  
  urlcolor=blue    
}

\begin{document}

\title{EVA-Net: Subject-Independent EEG Motor Decoding with Video-Derived Motor Priors}

\titlerunning{EVA-Net}

\author{
Ziyuan Li\textsuperscript{*} \and
Yueru Sun\textsuperscript{*} \and
Yimeng Zhang
}

\authorrunning{Z. Li, Y. Sun, Y. Zhang}

\institute{
South China University of Technology, Guangzhou, China\\
\email{\{202364870782,202364870212,202364870412\}@mail.scut.edu.cn}
}

\maketitle

\begingroup
\renewcommand{\thefootnote}{*}
\footnotetext[1]{These authors contributed equally to this work.}
\endgroup

\begin{abstract}
Practical non-invasive Brain-Computer Interface (BCI) systems require EEG decoders with strong cross-subject generalization and minimal calibration. However, inter-subject variability and signal non-stationarity often entangle motor semantics with subject-specific noise, limiting subject-independent decoding. Recent multimodal approaches use text as a semantic anchor, yet text provides sparse and static supervision for inherently dynamic motor processes. To address this issue, we propose EVA-Net, a two-stage framework that uses action videos as semantic priors for subject-independent EEG motor decoding. In the first stage, EEG and video features are aligned in a shared space using cross-modal and supervised contrastive objectives to reduce subject-specific variation. In the second stage, video category prototypes and knowledge distillation transfer video-derived priors to an EEG-only classifier without adding inference overhead. Experiments on two public datasets show that EVA-Net achieves strong subject-independent decoding performance, including an 8.66\% LOSO accuracy gain on EEGMMI. Ablation results further suggest that video provides a more effective semantic anchor than the text baseline considered in this work.

\end{abstract}

\section{Introduction}

Brain-computer interfaces (BCIs) are systems that establish direct communication channels with external devices by analyzing human brain electrical activity \cite{ref1}. Among these modalities, non-invasive electroencephalography (EEG) measures brain activity from scalp potentials and is widely used in motor imagery (MI) and motor execution (ME) tasks. Applications include stroke rehabilitation, virtual reality, gaming, and robotic arm control \cite{ref2,ref3,ref4}. Despite the success of deep learning in BCI applications \cite{ref5}, extracting robust, subject-independent representations remains a critical bottleneck. The inherent non-stationarity and inter-subject variability of EEG signals severely degrade cross-subject generalization, hindering the deployment of calibration-light BCI systems \cite{ref6,ref7,ref8}.

Recent multimodal paradigms, such as CLIP \cite{ref9}, demonstrate that contrastive learning across disparate modalities yields highly generalizable representations. While recent neuro-engineering studies attempt to utilize discrete text as semantic anchors to guide EEG representation learning, textual descriptions may not fully capture the continuous, fine-grained spatiotemporal dynamics of human motor execution. Conversely, action videos provide richer and more temporally informative motor priors. Aligning EEG signals with corresponding video sequences encourages the network to learn structured, high-level motion semantics and provides a potentially more informative way to reduce subject-specific variation.

Motivated by this, we propose \textbf{EVA-Net}, a subject-independent EEG motor decoding framework that leverages video-derived motor priors to facilitate calibration-light BCI deployment. EVA-Net adopts a two-stage training paradigm consisting of cross-modal alignment and prior-guided classification: 
1) Cross-Modal Alignment Stage: Initialized with pre-trained weights, a dual-encoder architecture—comprising a pre-trained CNN-Transformer hybrid EEG encoder and a pre-trained self-supervised video encoder—extracts modality-specific features. A dedicated alignment module then maps these features into a shared semantic space, utilizing contrastive learning to promote cross-modal semantic consistency between EEG signals and visual motion patterns.
2) Prior-Guided Classification Stage: To retain EEG-only inference at runtime, we introduce a video-prototype-based classifier coupled with knowledge distillation. This transfers robust video-derived motor priors to the EEG network without incurring real-time multimodal computational overhead. Key contributions are summarized as follows:

\begin{itemize}
\item We propose \textbf{EVA-Net}, an EEG--video framework for subject-independent motor decoding that uses video only during offline training and retains EEG-only inference.

\item We introduce a cross-modal alignment and prior-transfer scheme that preserves motor semantics while reducing subject-specific variation.

\item EVA-Net improves subject-independent decoding over the compared baselines in our experiments.
\end{itemize}

\section{Related Work}
\subsection{EEG Decoding}

In recent years, deep learning has emerged as the mainstream approach for analyzing EEG signals. Early CNN models, such as ShallowConvNet \cite{ref6} and EEGNet \cite{ref12}, leveraged their hierarchical feature extraction capabilities to capture EEG spatiotemporal patterns, achieving superior performance over traditional machine learning methods in both MI and ME tasks. However, CNNs are constrained by their local receptive fields, making it challenging to capture long-term temporal context dependencies. With the introduction of the Transformer architecture \cite{ref7}, its self-attention mechanism offered a novel approach for modeling global dependencies. Hybrid models like EEG Conformer \cite{ref13} emerged, combining convolutional modules with Transformer encoders to significantly enhance model performance in complex EEG tasks \cite{ref14}. These models demonstrated strong capabilities across multiple public EEG datasets (e.g., BCI Competition IV-2a/2b \cite{ref15,ref16}, and the SEED dataset \cite{ref17}). Although these unimodal approaches demonstrate strong performance in subject-dependent scenarios, their capabilities often deteriorate significantly when applied to unseen subjects due to substantial inter-individual variability in EEG signals. 

\subsection{Cross-Subject Decoding}

Achieving robust subject-independent decoding remains a critical bottleneck for practical BCI applications. Due to the highly non-stationary and individual-specific nature of EEG signals, models trained on one group of subjects often struggle to transfer directly to new users. To address this challenge, researchers have proposed various domain adaptation and transfer learning strategies \cite{ref18}. For instance, MSVTNet \cite{ref19} and EEGCCT \cite{ref8} enhance cross-subject robustness through multi-scale feature fusion and compact convolutional Transformer architectures, respectively. However, these purely supervised learning approaches typically rely on substantial labeled data and struggle to fully decouple task-independent individual noise, resulting in limited generalization performance. 

\subsection{Multimodal Alignment}
Multimodal representation learning, exemplified by CLIP \cite{ref9}, demonstrates powerful zero-shot transfer capabilities by mapping disparate modalities into a unified latent space. Recent attempts have adapted this paradigm to BCI-related EEG decoding by utilizing textual prompts as semantic anchors for EEG representation learning \cite{ref10}. However, from a systems perspective, textual descriptors are inherently static and discrete. They struggle to precisely encapsulate the continuous, time-varying dynamic variations intrinsic to EEG motor signals. In contrast, the video modality inherently encapsulates fine-grained kinematic trajectories and temporal evolutions. We hypothesize that action videos can serve as richer dynamic semantic anchors, enabling closer spatiotemporal alignment with EEG sequences and potentially improving cross-subject generalization.

\section{Method}

\begin{figure}[t]
  \centering
  \includegraphics[width=\textwidth]{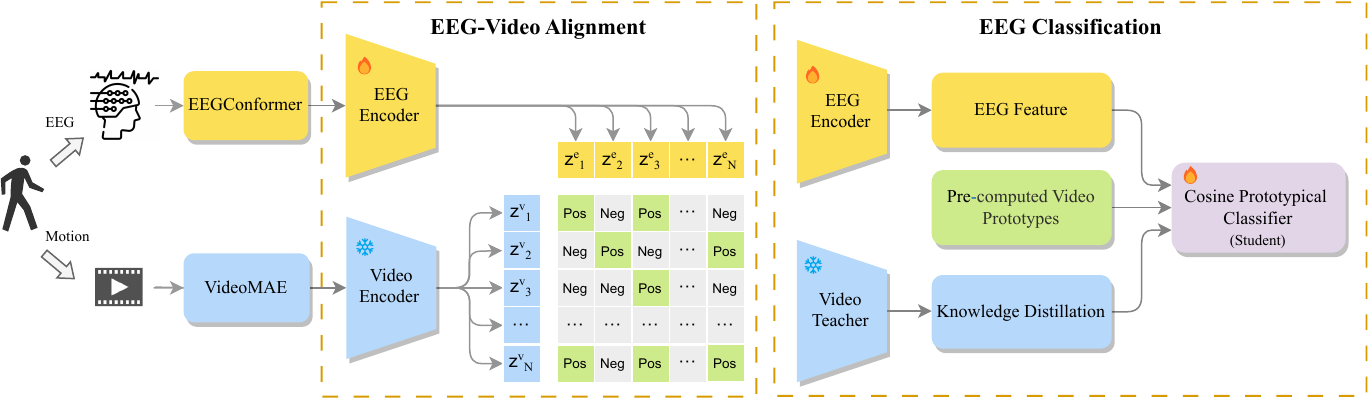}
  \caption{The overall framework of EVA-Net, consisting of an EEG--Video Alignment stage and an EEG Classification stage. $\mathbf{z}^{(e)}_i$ and $\mathbf{z}^{(v)}_i$ denote the shared-space embeddings of the $i$-th EEG trial and video sample, respectively, which are optimized using positive (Pos) and negative (Neg) pairs.}
  \label{fig1}
\end{figure}

The proposed EVA-Net consists of three core components: an EEG encoder, a video encoder, and a cross-modal alignment module (Fig.~\ref{fig1}). The pipeline operates through a hierarchical representation learning process. Initialized with pre-trained weights, the modality-specific encoders robustly extract high-dimensional feature representations from raw EEG signals and action videos. Subsequently, the alignment module utilizes projection heads to map and $L_2$-normalize these features into a shared semantic latent space. Finally, a joint contrastive optimization objective is applied to align the cross-modal embeddings, thereby encouraging more generalized and subject-invariant semantic representations.

Inspired by the EEG Conformer feature extraction method \cite{ref13}, EVA-Net's EEG encoder builds upon the EEG Conformer encoder architecture, comprising a convolutional module and a self-attention module.
The convolutional module takes raw two-dimensional EEG experimental data as input, applying temporal convolutional layers along the time dimension and spatial convolutional layers along the electrode channel dimension. An average pooling layer is then employed to suppress noise interference while enhancing generalization capabilities. These representations are mapped into query ($\mathbf{Q}$), key ($\mathbf{K}$), and value ($\mathbf{V}$) matrices to model global temporal dependencies using scaled dot-product self-attention:
\begin{equation}
\operatorname{Attention}(\mathbf{Q}, \mathbf{K}, \mathbf{V})
=
\operatorname{softmax}\!\left(
\frac{\mathbf{Q}\mathbf{K}^{\top}}{\sqrt{d_k}}
\right)\mathbf{V}.
\end{equation}
Here, $\mathbf{Q}, \mathbf{K} \in \mathbb{R}^{n \times d_k}$ and $\mathbf{V} \in \mathbb{R}^{n \times d_v}$ denote the query, key, and value matrices, respectively, and $d_k$ is the key dimension.

\subsection{Video Encoder Module}
We employ VideoMAE \cite{ref11}, a video self-supervised pretraining method based on high-ratio masking, to extract robust visual representations. Input video clips are partitioned into non-overlapping spatio-temporal tubelets, linearly projected into tokens, and combined with spatio-temporal positional embeddings. To capture global video-level semantics, a learnable class token (CLS) is prepended to the sequence. This entire sequence is then processed by the pre-trained VideoMAE encoder, and the resulting CLS vector is extracted as the global semantic representation of the video clip.

\subsection{Cross-Modal Contrastive Learning Framework}

Standard image-text contrastive frameworks are primarily optimized for static semantic matching, making them ill-equipped to handle the continuous spatiotemporal dynamics inherent in human motor control. To bridge the representational gap between non-stationary EEG time-series and kinetic video sequences, we propose a dual-component cross-modal framework. Beyond symmetric projection, our architecture explicitly introduces an inter-modal alignment phase and a subsequent visual-prior-guided classification phase, improving generalization to unseen subjects.

\subsubsection{Cross-Modal Alignment}
Given an EEG trial and its corresponding action video clip, encoders for modalities EEG and video extract feature vectors. Their outputs are mapped onto a shared space of the same dimension via a projection head and undergo L2 normalization \cite{ref20} to enhance training stability. 
We denote the projected and normalized EEG and video embeddings as \(\mathbf{z}^{(e)}_i\) and \(\mathbf{z}^{(v)}_j\in\mathbb{R}^{d}\), respectively, and define the cross-modal matching score directly as their cosine similarity $s_{ij}=\cos\!\left(\mathbf{z}^{(e)}_i,\mathbf{z}^{(v)}_j\right)$.

Since multiple EEG trials correspond to multiple video segments within a single category, for a given EEG sample, all video samples of the same category within the batch are considered positive instances, while all samples of other categories are negative instances. Bidirectional alignment is employed. The loss for an EEG sample \textit{i} can be expressed as:
\begin{equation}
\mathcal{L}_{e \rightarrow v}^{(i)}
=
- \log
\frac{\sum_{j:\, y_j = y_i}\exp\!\left(s_{ij}/\tau\right)}
{\sum_{k=1}^{B}\exp\!\left(s_{ik}/\tau\right)},
\tag{2}
\end{equation}
where \(\tau\) is a learnable temperature parameter, \(y_i\) is the class label of sample \textit{i}, and \(B\) is the batch size. The temperature parameter controls the sharpness of the softmax distribution.
Due to bidirectional alignment, the overall MIL-NCE loss is a symmetric average:
\begin{equation}
\mathcal{L}_{\mathrm{MIL}}
=
\frac{1}{2B}
\sum_{i=1}^{B}
\left(
\mathcal{L}_{e \rightarrow v}^{(i)}
+
\mathcal{L}_{v \rightarrow e}^{(i)}
\right),
\tag{3}
\end{equation}

To directly enhance intra-class compactness and inter-class separation in EEG representations, we incorporate a multi-positive supervised contrastive loss, inspired by supervised contrastive learning \cite{ref21}, within EEG data, making similar EEG embeddings closer and dissimilar ones farther apart. For an EEG anchor \(\mathbf{z}^{(e)}_i\), the loss is defined as:
\begin{equation}
\mathcal{L}_{\mathrm{sup}}^{(i)}
=
- \log
\frac{\sum_{p \in \mathcal{P}(i)} \exp\!\left(s^{(e)}_{ip}/\tau\right)}
{\sum_{k \ne i} \exp\!\left(s^{(e)}_{ik}/\tau\right)},
\tag{4}
\end{equation}
where \(\mathcal{P}(i)\) denotes the index set of all other positive samples (same category as \textit{i}) within the batch.
Thus, the overall loss function \(\mathcal{L}_{\mathrm{align}}\) for the alignment phase is a weighted sum of a cross-modal objective and an intra-modal objective:
\begin{equation}
\mathcal{L}_{\mathrm{align}}
=
\mathcal{L}_{\mathrm{MIL}}
+
\lambda_{\mathrm{sup}} \mathcal{L}_{\mathrm{sup}},
\tag{5}
\end{equation}
where \(\lambda_{\mathrm{sup}}\) is a hyperparameter balancing the two terms.

\subsubsection{Prior-Guided Classification}
Our classification module comprises an EEG encoder and a cosine prototype head \cite{ref22} that incorporates video-derived motor priors during offline training. Rather than relying on dynamic video pairings, it computes cosine similarities between EEG features and precomputed video category prototypes. This design helps reduce the influence of cross-subject and cross-scene variation in the video modality on the decision boundary, while introducing no real-time video processing overhead during inference. Specifically, for each category \(c\in\{1,\ldots,C\}\), we obtain $K$ prototypes by applying K-means clustering to all corresponding offline video embeddings.

During the forward propagation process, given an EEG embedding \(\mathbf{z}^{(e)}_i\), we first
compute its cosine similarities to all \(C \times K\) prototypes. Subsequently, the K similarities within the same category are aggregated using Log-Sum-Exp (LSE) to produce the logit for that category, thereby selecting the category with the highest score. This logit is then scaled using the learnable temperature \(\tau_c\):
\begin{equation}
g_c\!\left(\mathbf{z}^{(e)}_i\right)=\frac{1}{\tau_c}\log \sum_{k=1}^{K}\exp\!\left(\cos\left(\mathbf{z}^{(e)}_i, \mathbf{p}_{c,k}\right)\right),
\tag{6}
\end{equation}

LSE provides a smooth approximation to the max function and allows multiple prototypes within a category to contribute jointly. It dominates category scores when a 
prototype strongly matches a sample. When a sample aligns with multiple prototypes, it allows co-contribution from multiple centers, better accommodating diversity within the same category.

To further optimize the EEG embeddings \(\mathbf{z}^{(e)}_i\) and the learnable temperature \(\tau_c\), we employ three objectives. Label-smoothed cross-entropy \(\mathcal{L}_{\mathrm{CE}}\) provides the main supervision signal \cite{ref23}. We further introduce a response-based knowledge distillation loss \cite{ref24} to transfer class-level semantic information from the video modality. Specifically, a frozen video teacher produces \(C\)-class logits \(\boldsymbol{\ell}^{(t)}_i\) for each paired video input, while the EEG student produces logits \(\boldsymbol{\ell}^{(s)}_i\) for the corresponding EEG sample. Their temperature-scaled distributions are
\(
\mathbf{p}^{(t,T)}_i = \softmax(\boldsymbol{\ell}^{(t)}_i/T)\),  
\(\mathbf{p}^{(s,T)}_i = \softmax(\boldsymbol{\ell}^{(s)}_i/T)\), 
and the distillation loss is:
\begin{equation}
\mathcal{L}_{\mathrm{KD}}
=
\frac{T^{2}}{B}
\sum_{i=1}^{B}
\KL\!\left(
\mathbf{p}^{(t,T)}_i \,\middle\|\, \mathbf{p}^{(s,T)}_i
\right),
\tag{7}
\end{equation}
where \(B\) is the batch size and \(T\) is the distillation temperature. This term complements \(\mathcal{L}_{\mathrm{CE}}\) by transferring inter-class similarity information beyond one-hot labels.

To preserve alignment with the video-prototype space during classifier optimization, we further introduce an auxiliary prototype-logit regularization term \(\mathcal{L}_{\mathrm{proto}}\), which complements the main EEG-classifier cross-entropy by penalizing incorrect prototype-logit predictions computed from cosine similarities between EEG embeddings and video category prototypes. The total classification loss is:
\begin{equation}
\mathcal{L}_{\mathrm{cls}}
=
\alpha \mathcal{L}_{\mathrm{CE}}
+
\beta \mathcal{L}_{\mathrm{KD}}
+
\gamma \mathcal{L}_{\mathrm{proto}}.
\tag{8}
\end{equation}
Here, \(\alpha\), \(\beta\), and \(\gamma\) are hyperparameters that balance the relative contributions of supervised classification, knowledge transfer, and prototype-based regularization, respectively.

\section{Experiments}
\subsection{Setup}

\subsubsection{Datasets:}

We evaluate EVA-Net on two public datasets: \textbf{1) EEG Motor Movement/Imagery} (EEGMMI) \cite{ref25}: a 64-channel, 160 Hz EEG dataset from 109 subjects performing left/right/bilateral fist and bilateral foot movements across 14 runs. 
\textbf{2) BCI Competition IV Dataset 2a} (BCIC-IV-2a) \cite{ref15}: a cue-based motor imagery dataset containing 9 subjects, four classes (left hand, right hand, feet, and tongue), two sessions, and 288 trials per session. To establish cross-modal correspondence, we constructed
an auxiliary video dataset for category-level pairing. To cover the label spaces of both EEGMMI and BCIC-IV-2a, the video dataset included five action categories: left hand, right hand, bilateral hands, feet, and tongue. Ten subjects recorded these actions across five distinct scenarios to increase visual diversity. The continuous videos were segmented into 4-s clips, yielding 1000 clips per category.

\subsubsection{Baselines:}
We compare EVA-Net with five prominent deep learning baselines for EEG decoding: 1) EEGNet \cite{ref12}: a lightweight EEG-CNN, 2) ShallowConvNet \cite{ref6}:  a shallow EEG-CNN, 3) EEGCCT \cite{ref8}: a compact EEG transformer, 4) EEG Conformer \cite{ref13}: a CNN–Transformer fusion architecture, and 5) MSVTNet \cite{ref19}: a multi-scale transformer for MI-EEG decoding.

\subsubsection{Protocols:}

To evaluate the performance of the proposed model as a generalizable EEG motion decoding algorithm and mitigate overfitting risks from limited single-subject samples, we designed three experimental categories:
\textbf{1) Cross-Session Subject-Dependent}: 
session 1 is used for training and session 2 for testing. We apply 5-fold cross-validation on the training set to create a validation split for early stopping. 
\textbf{2) Subject-Independent} (LOSO): 
we adopt the Leave-One-Subject-Out (LOSO) cross-validation protocol, in which each subject is held out as the test set once, while the data from all remaining subjects are used for training.
\textbf{3) Pooled-Subject K-Fold Cross-Validation}: 
all samples across subjects are randomly pooled before performing K-fold cross-validation. This strategy specifically mitigates the overfitting risks caused by the limited single-subject data in the EEGMMI dataset.

\subsubsection{Implementation Details:}

EVA-Net is implemented in PyTorch on NVIDIA A100 GPUs. The EEG Conformer-based encoder was pretrained separately within each experimental split using only the corresponding training data. No validation or test samples, including the held-out subject in LOSO evaluation, were used during pretraining or subsequent model training. It outputs 256-D representations (dropout 0.5). For the video modality, we employ a VideoMAE model pre-trained on the Kinetics-400 dataset \cite{ref26}. It processes 16-frame, $224 \times 224$ crops into 384-D features, which are then mapped to the 256-D shared space via a 2-layer MLP projection head (dropout 0.3).

Stage 1: Cross-Modal Alignment. To preserve stable semantic anchors, the VideoMAE backbone is kept frozen. Only the video projection head and the last two EEG Transformer blocks are trained via class-balanced sampling. For $\mathcal{L}_{\mathrm{align}}$, we set $\lambda_{\text{sup}} = 0.15$ and initialize $\tau=0.05$. We optimize using AdamW (weight decay $10^{-3}$) with a 5-epoch linear warmup and cosine decay. Learning rates are $2 \times 10^{-3}$ (projection head), $1 \times 10^{-3}$ (EEG encoder), and $1 \times 10^{-4}$ ($\tau$). Mixed precision and early stopping are applied.

Stage 2: Prior-Guided Classification. With the video modality kept frozen, we pre-compute $K=2$ prototypes per category to initialize the classification head. We fine-tune only the EEG classification head and its final Transformer block for 40 epochs. For $\mathcal{L}_{\mathrm{cls}}$, we apply a label smoothing of 0.05, $\alpha=1.0$, $\beta=0.07$, $T=2.0$, and $\gamma=0.15$. Learning rates are $8 \times 10^{-4}$ (classification head) and $2 \times 10^{-4}$ (EEG encoder), updated with EMA (decay 0.997).

\subsection{Main Results}
As shown in Table~\ref{tab1}, EVA-Net delivers its most substantial improvements under the Leave-One-Subject-Out (LOSO) protocol, which is the most practically relevant setting for calibration-light BCI deployment because it directly evaluates generalization to unseen subjects. On EEGMMI, EVA-Net achieves 72.30\% Accuracy and 70.80\% Macro-F1 in the LOSO setting, outperforming the strongest compared baseline by 8.66 and 8.13 percentage points, respectively. It also attains the highest Kappa score (0.70), exceeding the best baseline by 0.13. These gains are notably larger than those observed under pooled-subject K-fold cross-validation, where EVA-Net still ranks first but with more moderate improvements. This pattern suggests that the proposed video-guided semantic alignment is particularly beneficial when subject-specific variability becomes the main obstacle to decoding performance.

A similar trend is observed on the BCIC-IV-2a dataset in Table~\ref{tab2}. Although EVA-Net does not achieve the best result in the subject-dependent setting, it yields the highest accuracy in the subject-independent evaluation, reaching 71.25\% and surpassing the strongest baseline (EEGCCT) by 3.35 percentage points. Notably, EVA-Net improves the subject-independent result while remaining competitive in the subject-dependent setting, indicating that its gains do not arise merely from overfitting to subject-specific patterns. Taken together, the results on both datasets show that EVA-Net is particularly effective in subject-independent scenarios, where robust cross-subject transfer is critical.

\begin{table}[t]
\caption{Comparison of decoding performance (mean $\pm$ std) on the EEGMMI dataset}
\label{tab1}
\centering
\small
\setlength{\tabcolsep}{4.5pt} 

\makebox[\textwidth][c]{%
\begin{tabular}{c l c c c}
\toprule
Experimental Setup & Methods & Accuracy & Macro-F1 & Kappa \\
\midrule
\multirow{5}{*}{\parbox{3.4cm}{\centering pooled-subject K-fold\\cross-validation}}
 & EEGNet~\cite{ref12}        & 67.30\%$\pm$3.12\% & 66.11\%$\pm$3.21\% & 0.62$\pm$0.03 \\
 & ShallowConvNet~\cite{ref6} & 60.72\%$\pm$1.02\% & 60.54\%$\pm$1.16\% & 0.48$\pm$0.01 \\
 & EEG Conformer~\cite{ref13} & 72.69\%$\pm$2.13\% & 72.74\%$\pm$1.99\% & 0.64$\pm$0.03 \\
 & MSVTNet~\cite{ref19}       & 66.83\%$\pm$2.13\% & 66.66\%$\pm$2.09\% & 0.56$\pm$0.20 \\
 & \textbf{EVA-Net}           & \textbf{76.10\%$\pm$2.47\%} & \textbf{76.05\%$\pm$2.42\%} & \textbf{0.66$\pm$0.03} \\
\midrule
\multirow{6}{*}{\parbox{3.4cm}{\centering Leave-One-Subject-Out}}
 & EEGNet~\cite{ref12}        & 63.64\%$\pm$5.61\% & 62.67\%$\pm$6.53\% & 0.51$\pm$0.14 \\
 & ShallowConvNet~\cite{ref6} & 60.79\%$\pm$7.62\% & 60.80\%$\pm$7.91\% & 0.57$\pm$0.15 \\
 & EEGCCT~\cite{ref8}         & 63.34\%$\pm$8.51\% & --                 & -- \\
 & EEG Conformer~\cite{ref13} & 62.56\%$\pm$8.61\% & 61.40\%$\pm$9.60\%  & 0.50$\pm$0.12 \\
 & MSVTNet~\cite{ref19}       & 53.00\%$\pm$9.18\% & 55.20\%$\pm$12.00\%   & 0.42$\pm$0.15 \\
 & \textbf{EVA-Net}           & \textbf{72.30\%$\pm$8.90\%}   & \textbf{70.80\%$\pm$9.30\%}   & \textbf{0.70$\pm$0.12} \\
\bottomrule
\end{tabular}%
}
\end{table}

\begin{table}[!htbp]
\caption{Comparison of accuracy (mean $\pm$ std) on the BCIC-IV-2a dataset}
\label{tab2}
\centering
\small
\setlength{\tabcolsep}{4pt} 
\begin{tabular}{l c c}
\toprule
Methods &
\begin{tabular}[c]{@{}c@{}}subject-dependent\\ (session-independent)\end{tabular} &
subject-independent \\
\midrule
EEGNet~\cite{ref12}        & 71.52\%$\pm$11.33\% & 63.44\%$\pm$12.69\% \\
ShallowConvNet~\cite{ref6} & 69.78\%$\pm$9.00\%  & 59.43\%$\pm$12.46\% \\
EEGCCT~\cite{ref8}         & --                  & 67.90\%$\pm$12.65\% \\
EEG Conformer~\cite{ref13} & 69.28\%$\pm$11.80\% & 60.70\%$\pm$17.67\% \\
MSVTNet~\cite{ref19}       & \textbf{78.50\%$\pm$10.86\%} & 64.10\%$\pm$14.47\% \\
\textbf{EVA-Net}           & 75.80\%$\pm$7.90\% & \textbf{71.25\%$\pm$12.34\%} \\
\bottomrule
\end{tabular}
\end{table}

\begin{figure}[!htbp]
  \centering
  \includegraphics[width=0.9\textwidth]{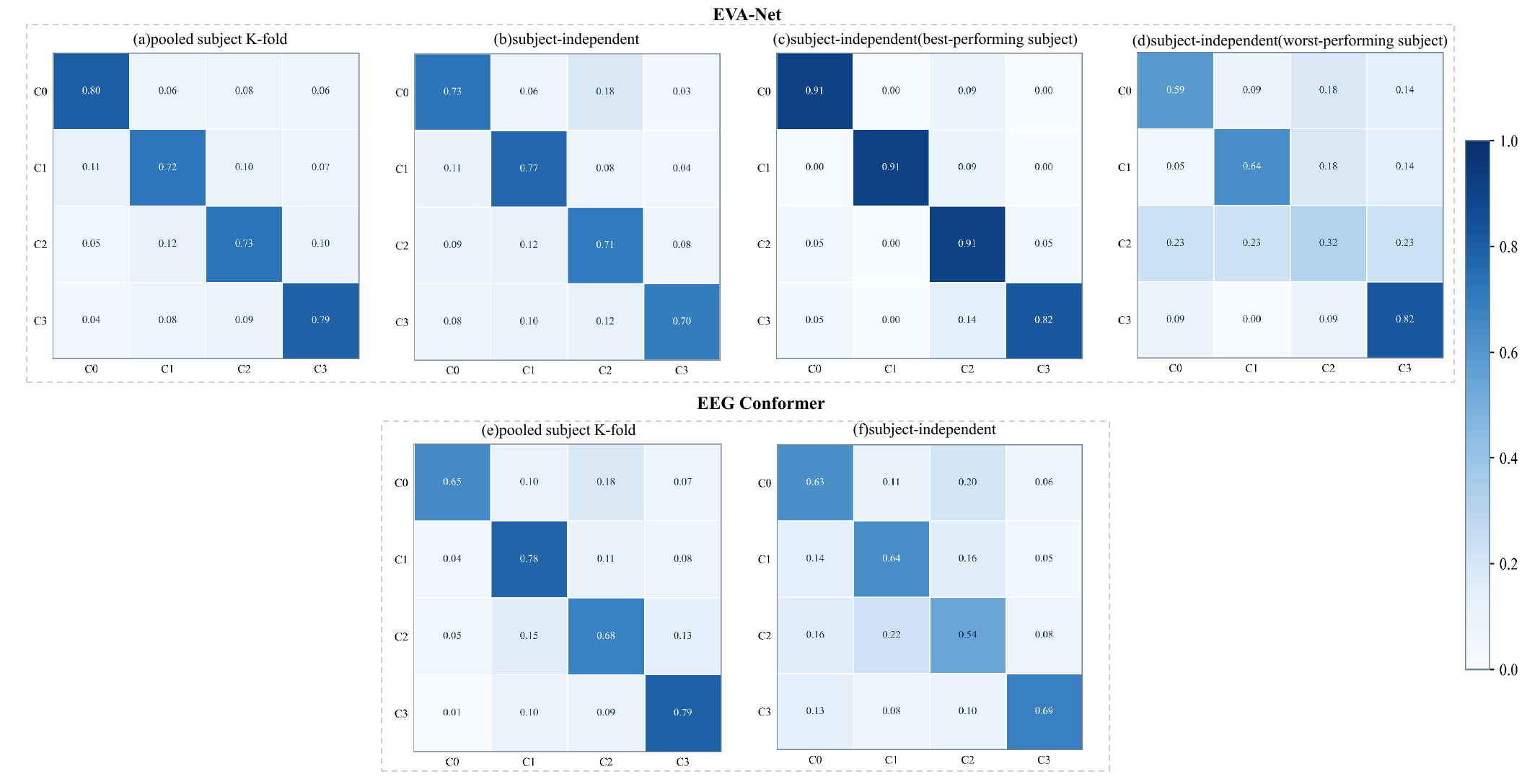}
  \caption{Normalized confusion matrices for different models on the EEGMMI dataset. Each row represents the true label and each column represents the predicted label.}
  \label{fig2}
\end{figure}

On the EEGMMI dataset, Fig.~\ref{fig2} presents row-normalized confusion matrices under both pooled K-fold and subject-independent settings. Compared with EEG Conformer, EVA-Net produces more concentrated diagonals and less off-diagonal confusion in both settings, suggesting improved class separability. Its advantage is especially evident under subject-independent evaluation, where higher category-wise recall is still maintained, suggesting improved cross-subject robustness. In addition, the best-performing subject is close to an ideal diagonal matrix, showing that EVA-Net can learn highly discriminative motion representations for favorable subjects. Nevertheless, the worst-performing subject still exhibits noticeable confusion in certain categories, implying that residual subject-specific variability and EEG non-stationarity remain challenging under LOSO evaluation.

\subsection{Ablation Study}

\begin{figure}
  \centering
  \includegraphics[width=0.8\textwidth]{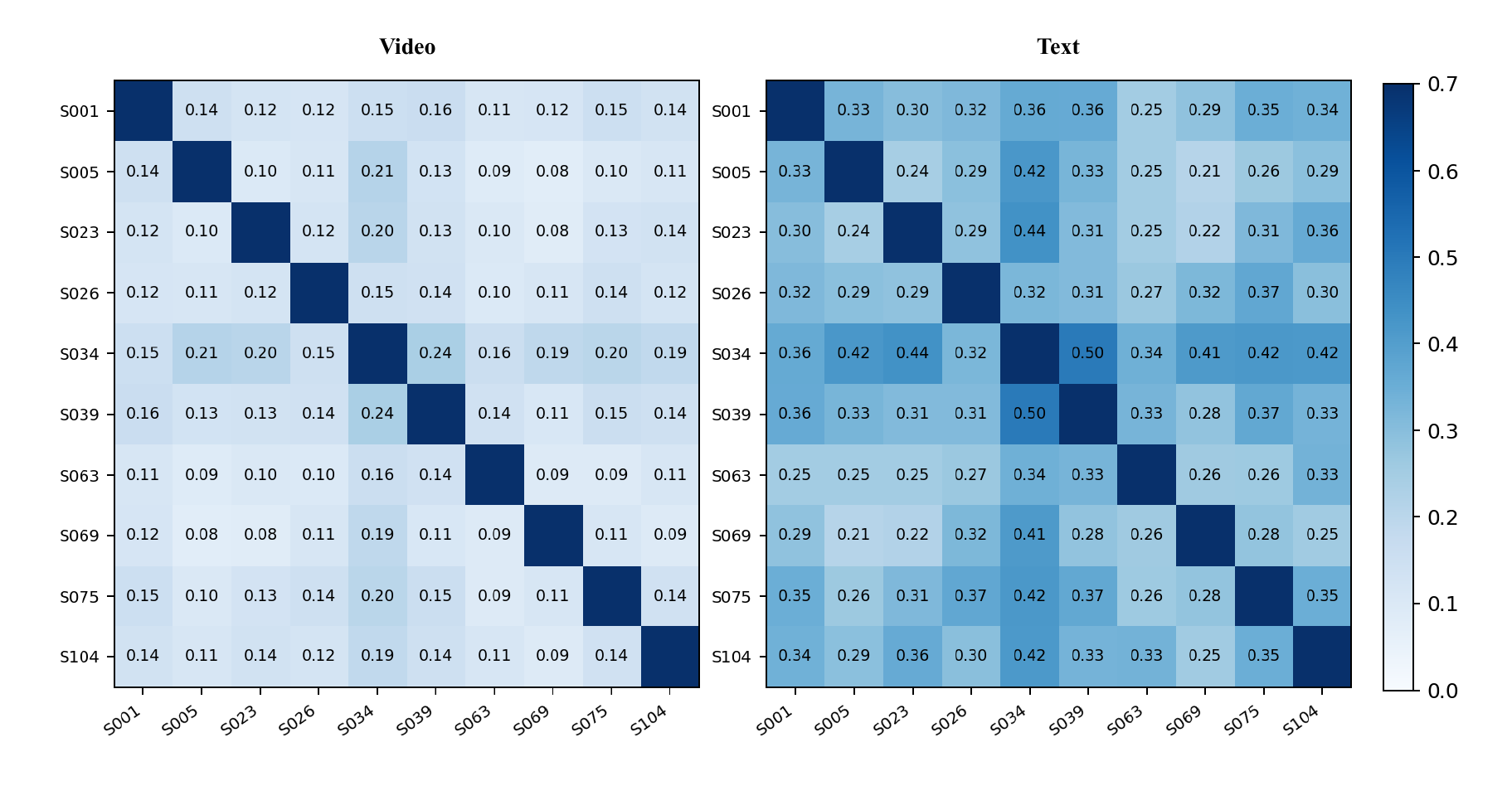}
  \caption{Heatmaps of cross-subject distributions under the video and text modalities. Both axes denote subject IDs (10 randomly selected subjects).}
  \label{fig3}
\end{figure}

To evaluate the benefit of the video modality, we replaced it with the text modality and retrained the model on the EEGMMI dataset under identical settings. The text encoder was a standard Transformer encoder. As shown in Fig.~\ref{fig3}, the cross-subject FID heatmaps show that text prototypes have greater variance across subjects, indicating that they are less effective at reducing inter-subject variability. These results suggest that, compared with the text modality, the video modality can serve as a more stable semantic anchor for EEG representation alignment.

\begin{figure}
  \centering
  \includegraphics[width=0.8\textwidth]{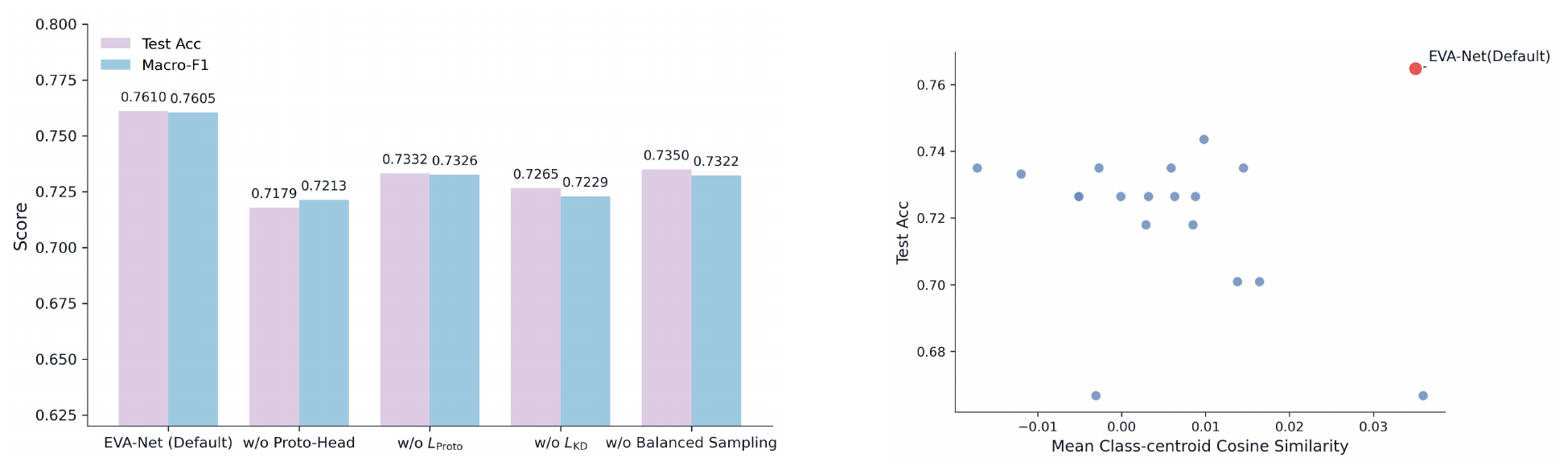}
  \caption{Ablation study of EVA-Net. Left: Component-wise ablation results on Test Acc and Macro-F1. Right: Accuracy vs. mean class-centroid cosine similarity across ablation variants.}
  \label{fig4}
\end{figure}
Beyond modality selection, we conducted a component-wise ablation study to evaluate EVA-Net's architectural designs (Fig.~\ref{fig4}). As shown in Fig.~\ref{fig4} (left), removing the prototype classification head (w/o Proto-Head) causes the most severe performance drop, highlighting the importance of video semantic priors in shaping the decision boundary. Other components ($\mathcal{L}_{\text{proto}}$, $\mathcal{L}_{\mathrm{KD}}$, and class-balanced sampling) also provide stable gains. Furthermore, Fig.~\ref{fig4} (right) reveals a strong positive correlation between classification accuracy and mean cross-modal cosine similarity. Our default model lies in the upper-right region, suggesting that stronger cross-modal semantic alignment is associated with better decoding performance.

\section{Conclusion}

This paper proposes EVA-Net, a framework for subject-independent EEG motor decoding.  EVA-Net uses action videos during offline training to provide structured video-derived motor priors, while retaining EEG-only inference at runtime. Specifically, the framework first aligns EEG trials and video embeddings in a shared semantic space to encourage subject-invariant motor representations, and then transfers these priors to an EEG classifier through prototype-guided classification and knowledge distillation. Consequently, it shows improved transferability under cross-subject shifts. Extensive experiments were conducted on two public datasets. Results show that EVA-Net achieves strong decoding performance in subject-independent evaluations and helps reduce overfitting to subject-specific patterns.

\end{document}